\def\Year{2000}

\documentstyle[coin,twoside]{article}
\include{mssymb} 

\def\None#1{}

\def\Small#1{}

\def\OK#1{}

\def\Noun#1{{\sf Noun #1}}
\def\Verb#1{{\sf Verb #1}}

\begin{document}

\title{Metonymy Interpretation Using X {\it no} Y Examples}

\maketitle

\author{Masaki Murata$^{*}$ 
\hspace{0.2cm} Qing Ma$^{*}$ 
\hspace{0.2cm} Atsumu Yamamoto$^{**}$ 
\hspace{0.2cm} Hitoshi Isahara$^{*}$}

\affil{  $^{*}$Communications Research Laboratory, 
  Ministry of Posts and Telecommunications\\
  588-2, Iwaoka, Nishi-ku, Kobe, 651-2492 Japan\\
  $^{**}$Kyoto University, Yoshida-Honmachi, Sakyo, 
  Kyoto, 606-8501 Japan
}

\begin{abstract}
We developed on example-based method 
of metonymy interpretation. 
One advantages of this method 
is that a hand-built database of metonymy is not necessary 
because it instead uses examples in the form ``Noun X {\it no} Noun Y 
(Noun Y of Noun X).'' 
Another advantage is that 
we will be able to interpret newly-coined metonymic sentences  
by using a new corpus. 
We experimented with metonymy interpretation 
and obtained 
a precision rate of 66\% when using this method. 
\kw{metonymy, ellipsis, example-based method, corpus}
\end{abstract}

\section{Introduction}\label{chap:intro}

This paper describes 
a new Japanese metonymy interpretation method 
using the example-based method 
(Nagao 1984; Murata and Nagao 1997; 
Murata et al. 1999b; Murata et al. 1999a). 
Metonymy is a metaphorical expression 
in which the name of something is substituted for another thing 
associated with the thing named. 
For example, in the Japanese sentence of 
``{\it boku ga torusutoi wo yomu} (I read Tolstoi),'' 
the word ``{\it torusutoi} (Tolstoi)'' is a metonymic word. 
In this case, 
the word ``{\it torusutoi} (Tolstoi)'' means 
a book written by him ``{\it torusutoi no hon} (Tolstoi's book).'' 
The actual meaning ``{\it torusutoi no hon} (Tolstoi's book)'' 
is replaced 
by the abbreviated expression ``{\it torusutoi} (Tolstoi).'' 
In this paper 
we call an abbreviated word such as ``{\it torusutoi} (Tolstoi)'' 
{\it a source word}, and  
call the word it replaces --- in this case, 
``{\it torusutoi no hon} (Tolstoi's book)'' --- {\it a target word}. 

Metonymy has conventionally 
has been interpreted by using a hand-built database 
that includes relationships between source words and 
target words, such as a special knowledge base of metonymy, 
and a semantic network (Iverson and Helmreich 1992; Fass 1988). 
Relationships between source words and target words are diverse, 
however, and it is not easy to make a detailed database. 
This paper 
therefore 
describes a method that instead 
interprets metonymy 
by using examples in the form of noun phrases 
such as ``\Noun{X} {\it no} \Noun{Y} (\Noun{Y} of \Noun{X})'' and 
``\Noun{X} \Noun{Y}.'' 
When we interpret a source word ``{\it torusutoi} (Tolstoi)'' 
in the sentence ``{\it boku ga torusutoi wo yomu} (I read Tolstoi),'' 
for example, 
we gather two sets of nouns from examples in the form of 
``{\it torusutoi-no} \Noun{Y} (\Noun{Y} of Tolstoi)'' 
and examples in the form of 
``\Noun{Z} {\it wo yomu} (read \Noun{Z}).'' 
We select a noun ``{\it shousetsu} (novel)'' 
included in both sets, \Noun{Y} and \Noun{Z}, 
and judge that 
the source word ``{\it torusutoi} (Tolstoi)'' means 
the target word ``{\it torusutoi no shousetsu} (Tolstoi's novel).'' 
Two advantages of this method are 
that a special knowledge of metonymy is not necessary 
because we use examples and 
that 
we can interpret newly-coined metonymic sentences 
by using a new corpus. 

Most metonymies are considered elliptical expressions, 
such that 
``I read Tolstoi'' is considered 
the elliptical expression of the literal sentence 
``I read Tolstoi's novel.'' 
We have developed an example-based method 
for ellipsis resolution and 
in previous work have used it in the resolution of 
indirect anaphora (Murata et al. 1999b), 
in pronoun resolution (Murata et al. 1999a) and 
in the resolution of verb ellipsis (Murata and Nagao 1997). 
In that work, 
examples were used as the semantic restriction. 
Consider the following example of 
indirect anaphora resolution (Murata et al. 1999b). 
``{\it I went into \underline{an old house} last night. 
\underline{The roof} was leaking badly and ...}'' 
indicates that ``{\it the roof}'' is associated with ``{\it an old house}.'' 
In this case 
we used ``the roof of a house'' in the form of 
``{\sf Noun X} of {\sf Noun Y}'' as examples and 
restricted the antecedent of ``{\it the roof}'' 
to ``{\it a house}.'' 
The metonymy interpretation method 
described in the present paper is very 
similar to that indirect anaphora resolution method 
in that it uses 
the examples in the form of ``{\sf Noun X} of {\sf Noun Y}'' 
as the semantic restiction. 
A valuable contribution of this paper is that 
it shows that 
metonymies can be resolved by the method used in 
ellipsis resolution. 

\section{Metonymy}\label{chap:riron}

There are many kinds of metonymic relationships,  
such as ``author-book'' and ``container-content,'' 
and 
some are listed and illustrated in Table \ref{meto}. 

We explain metonymy using the following example. 

{\small
\begin{tabular}[h]{lllll}
{\it boku} & {\it ga} & {\it torusutoi} & {\it wo} & {\it yomu.} \\
(I) & {\sf subj-case} & (Tolstoi) & {\sf obj-case} & (read) \\
\multicolumn{5}{l}{(I read Tolstoi.)}\\
\end{tabular}
}

This is a metonymic sentence 
using an ``author-book'' relationship. 
If we interpret the sentence  
``{\it boku ga torusutoi wo yomu} (I read Tolstoi)'' literally, 
it ridiculously means that one reads ``torusutoi'' himself. 
The word `` {\it torusutoi} (Tolstoi),'' 
which is an author's name, 
denotes 
a novel written by him. 
This sentence means 
``{\it boku ga torusutoi no hon wo yomu} (I read Tolstoi's book)''. 

\begin{table*}[t]
\footnotesize
\caption{Metonymic Relationships}
\label{meto}
\begin{tabular}{@{ }l@{ }|@{}l@{}|@{}l@{}}
\multicolumn{1}{c}{Relationship}&
\multicolumn{1}{|c}{Metonymic sentence}&
\multicolumn{1}{|c}{Interpretation}\\
\multicolumn{1}{c}{}&
\multicolumn{1}{|c}{(the source word)}&
\multicolumn{1}{|c}{(the target word)}\\
\hline
Author-book & 
\begin{tabular}[h]{l@{ }l@{ }l@{ }l@{ }l}
{\it boku} & {\it ga} & {\it \underline{torusutoi}} & {\it wo} & {\it yomu.}\\
(I)  & {\sf subj} & (Tolstoi) & {\sf obj} & (read)\\
\multicolumn{5}{l}{(I read \underline{Tolstoi})}\\
\end{tabular}
&
\begin{tabular}[h]{l@{ }l@{ }l}
{\it torusutoi} & {\it no} & {\it shousetsu.}\\
(Tolstoi)  & (of) & (a novel) \\
\multicolumn{3}{l}{(Tolstoi's novel)}\\
\end{tabular}
\\[0.5cm]\hline
Maker-product & 
\begin{tabular}[h]{l@{ }l@{ }l@{ }l@{ }l}
{\it boku} & {\it ga} & {\it \underline{ferari}} &  {\it ni} & {\it noru} \\
(I) & {\sf subj} & (Ferrari) &  {\sf obj} & (drive) \\
\multicolumn{5}{l}{(I drive a \underline{Ferrari}.)}\\
\end{tabular}
&
\begin{tabular}[h]{l@{ }l@{ }l@{ }l@{ }l}
{\it ferari} &  {\it no} & {\it kuruma} \\
(Ferrari) &  (of) & (a car) \\
\multicolumn{5}{l}{(a car made by Ferrari)}\\
\end{tabular}
\\[0.5cm]\hline
Container-content & 
\begin{tabular}[h]{l@{ }l@{ }l@{ }l@{ }l}
{\it boku} & {\it ga} & {\it \underline{nabe}} &  {\it wo} & {\it taberu} \\
(I) & {\sf subj} & (a pot) &  {\sf obj} & (eat) \\
\multicolumn{5}{l}{(I eat \underline{a pot}.)}\\
\end{tabular}
&
\begin{tabular}[h]{l@{ }l}
{\it nabe} &  {\it ryori} \\
(a pot) &  (food) \\
\multicolumn{2}{l}{(hot-pot food)}\\
\end{tabular}
\\[0.5cm]\hline
Agent-attachment & 
\begin{tabular}[h]{l@{ }l@{ }l}
{\it \underline{tsume-eri}} & {\it ga} & {\it aruite kuru} \\
(A stand-up collar) & {\sf subj} & (come walking) \\
\multicolumn{3}{l}{(\underline{A stand-up collar} comes walking.)}\\
\end{tabular}
&
\begin{tabular}[h]{l@{ }l@{ }l}
{\it tsume-eri} & {\it no} & {\it gakusei} \\
(A stand-up collar) & (of) & (a student) \\
\multicolumn{3}{l}{(a student with a stand-up collar)}\\
\end{tabular}
\\[0.5cm]\hline
Agent-neighbor & 
\begin{tabular}[h]{l@{ }l@{ }l@{ }l@{ }l}
{\it \underline{hamusando}} & {\it ga} & {\it kanjo} & {\it wo} & {\it harau} \\
(a ham sandwich) & {\sf subj} & (a bill) & {\sf obj} & (pay) \\
\multicolumn{5}{l}{(\underline{A ham sandwich} pays a bill.)}\\
\end{tabular}
&
\begin{tabular}[h]{l@{ }l@{ }l}
{\it hamusando} & {\it no} & {\it kyaku} \\
(a ham sandwich) & (of) & (a customer) \\
\multicolumn{3}{l}{(a customer eating a ham sandwich)}\\
\end{tabular}
\\[0.5cm]\hline
\None{
Placename-product & 
\begin{tabular}[h]{l@{ }l@{ }l@{ }l@{ }l}
{\it boku} & {\it ga} & {\it \underline{oshima}} & {\it wo} & {\it shitateru} \\
(I) & {\sf subj} & {\footnotesize (Oshima)} & {\sf obj} & {\footnotesize (tailor)} \\
\multicolumn{5}{l}{(I tailor \underline{Oshima}.)}\\
\end{tabular}
&
\begin{tabular}[h]{l@{ }l}
{\it {oshima}} & {\it tsumugi} \\
{\footnotesize (Oshima)} & (pongee) \\
\multicolumn{2}{l}{(a Oshima pongee.)}\\
\end{tabular}
\\[0.5cm]\hline
}
Place-thing & 
\begin{tabular}[h]{l@{ }l@{ }l@{ }l@{ }l}
{\it boku} & {\it ga} & {\it \underline{niwa}} & {\it wo} & {\it haku.} \\
(I) & {\sf subj} & {\footnotesize (a garden)} & {\sf obj} & {\footnotesize (sweep)} \\
\multicolumn{5}{l}{(I sweep \underline{a garden}.)}\\
\end{tabular}
&
\begin{tabular}[h]{l@{ }l@{ }l}
{\it niwa} & {\it no} & {\it gomi} \\
(a garden) & (of) & (a rubbish) \\
\multicolumn{3}{l}{(the rubbish in a garden)}\\
\end{tabular}
\\[0.5cm]\hline
\end{tabular}
\end{table*}

In the work described in this paper 
we dealt only with metonymy, not with metaphor. 
The reason is as follows. 
Metaphor is affected by the context, 
so metaphor interpretation is difficult. 
But metonymy, which is based on the associative relationship between words, 
can be interpreted easily by using only 
simple information such as the relationships between words. 

\section{Example-Based Metonymy Interpretation}\label{chap:system}


\subsection{Metonymy and Examples in the form of ``\Noun{X} of \Noun{Y}''}

Metonymy is a writing technique based on replacement.  
In the sentence ``{\it boku ga torusutoi wo yomu} (I read Tolstoi)'' 
the target word ``{\it shousetsu} (a novel)'' is replaced 
by the source word ``{\it torusutoi} (Tolstoi)''. 
But it is not simple replacement. 
``{\it shousetsu} (a novel)'' does not generically refer to 
all novels, 
but to a certain novel ``{\it torusutoi no shousetsu} (Tolstoi's novel).'' 
It can be thought that 
in ``{\it boku ga torusutoi wo yomu} (I read Tolstoi)'', 
the relationship of ``author-book'' is recognized 
by unspoken agreement 
and that ``{\it shousetsu} (a novel)'' in 
`` {\it torusutoi no shousetsu} (Tolstoi's novel)'' is omitted.  

Many metonymic sentences could thus be interpreted 
by inferring an elliptical target word having a certain relationship 
to a source word. 
Because source and target words in metonymy have 
a relationship, such as the ``author-book'' relationship, 
we can search for the target word by using the source word. 

This paper, thus describes a method by which 
metonymy is interpreted by 
using the relationships between words to infer an elliptical word. 
We search for these relationships by using 
examples, and the ones 
we use are of the following kinds:\footnote{When we use 
our method in English, 
we should use not only 
``\Noun{X} of \Noun{Y}'' and ``\Noun{X}'s \Noun{Y}'' but also 
``\Noun{X} in \Noun{Y},'' 
``\Noun{X} on \Noun{Y},'' 
``\Noun{X} for \Noun{Y}'' and so on.}. 
\begin{itemize}
\item ``\Noun{X} {\it no} \Noun{Y} (\Noun{Y} of \Noun{X})''
\item ``\Noun{X} \Noun{Y} (\Noun{X} \Noun{Y})''
\end{itemize}
The relationships between \Noun{X} and \Noun{Y} in ``\Noun{X} {\it no} \Noun{Y}'' 
include almost all the relationships of metonymy, 
so examples in the form of ``\Noun{X} {\it no} \Noun{Y}'' 
can be used in metonymy interpretation. 
(Note that the Japanese particle ``{\it no}'' has many meanings. 
It is semantically similar 
to the English preposition ``of'' but 
has many more meanings. 
It can, for example, express the meanings of 
``at,'' ``in,'' ``for,'' ``with,'' ``by,'' and 
many other noun-noun relationships.) 
Metonymy is interpreted by 
gathering as examples 
noun phrases in the form of ``\Noun{X} {\it no} \Noun{Y},'' 
selecting the most appropriate word among their \Noun{Y}s, 
and substituting it in the metonymy sentence. 
In some cases, however, 
we make noun phrases without using the particle ``{\it no}.'' 
The phrase 
``{\it nabe no ryori} (food of a pot),'' for example, is not used 
but instead the phrase ``{\it nabe ryori} (pot food)'' is generally used, 
as in 
the metonymic sentence ``{\it boku ga \underline{nabe} wo taberu} 
(I eat \underline{a pot}),'' 
which means ``{\it boku ga \underline{nabe ryori} wo taberu} 
(I eat \underline{pot food}).'' 
In order to deal with such cases, 
we also use compound nouns ``\Noun{X} \Noun{Y}'' as examples 
in metonymy interpretation. 

\subsection{Example-Based Metonymy Interpretation}

\None{
In our method, 
metonymy interpretation is done by gathering examples and using them. 
}

Metonymy detection cannot be separated from metaphor interpretation 
because 
after we detect non-literal sentences, 
we must distinguish metonymic sentences from metaphorical sentences. 
But as mentioned in Section \ref{chap:riron}, 
we did not handle metaphor interpretation in the work reported here. 
Therefore we did not handle metonymy detection 
and simply assumed that 
the input sentences were 
metonymic sentences. 

Metonymy interpretation was performed as follows. 
\begin{enumerate}
\item We specified a source word (``\Noun{X}'') by 
using a case frame dictionary. 
\item 
We gathered from a corpus examples 
in the forms of ``\Noun{X} {\it no} \Noun{Y} (\Noun{Y} of \Noun{X})'' and ``\Noun{X} \Noun{Y}'' 
that included the source word ``\Noun{X}'' 
and we got as candidate target words many \Noun{Y}s. 
\item 
We selected one of these candidates as the target word 
by using 
semantic restriction of the verb and using the frequency of \Noun{Y}. 
\end{enumerate}

\subsubsection{Specification of the source word}

Our method uses a case frame dictionary to specify the source word. 
An input sentence is transformed 
by a morphological analyzer and a syntactic analyzer
into the following structure. 
{\small
\begin{quote}
``Noun $+$ Case-Particle'', \, ``Noun $+$ Case-Particle'', \, $\cdots,$ \, ``Predicative-Verb''  
\end{quote}
}
We compare such a structure to a case frame of the verb, 
and check whether or not 
each of the nouns satisfies the semantic restriction 
in the case frame dictionary. 
We judge that 
the source word is 
the noun that does not satisfy the semantic restriction. 

Let us specify 
the source word in the following metonymic sentence. 

{\small
\begin{tabular}[h]{lllll}
{\it boku} & {\it ga} & {\it torusutoi} & {\it wo} & {\it yomu.}\\
(I)  & {\sf subj-case} & (Tolstoi) & {\sf obj-case} & (read)\\
\multicolumn{5}{l}{(I read Tolstoi.)}\\
\end{tabular}
}

Suppose that the case frame of ``{\it yomu} (read)'' is as follows. 

{\small
\begin{quote}
\{human\} {\sf subj-case particle}, \, \{book, newspaper, novel\} {\sf obj-case particle}, \, \{read\}
\end{quote}
}

We check the semantic satisfaction of each case element 
by comparing the input and the case frame. 
Because ``{\it boku} (I)'' in the subjective-case of 
the input is a human being, 
this word satisfies the semantic restriction. 
``{\it torusutoi} (Tolstoi)'' 
in the objective-case does not belong to 
the book, newspaper, or novel category, 
so it does not satisfy the semantic restriction. 
Therefore, we judge that 
``{\it torusutoi} (Tolstoi)'' 
in the objective-case is the source target. 

\subsubsection{Extraction of Target Word Candidates}
\label{tyuusyutu}

We gather from a corpus 
sentences including 
the source word (``\Noun{X}'') and transform them into syntactically 
tagged data by using a syntactic analyzer. 
We consider as examples containing candidate target words 
examples in the form of 
``\Noun{X} {\it no} \Noun{Y}'' and ``\Noun{X} \Noun{Y}.'' 
We consider the frequency of anay candidate target word to be 
the number of examples containing that word. 

We also gather from the corpus sentences including 
the verb (``\Verb{W}'') acting on \Noun{X} 
in the metonymic sentence 
and transform them into syntactic structures 
by syntactic analysis. 
We extract examples in the form of ``\Noun{Z} \, Case-Particle \, \Verb{W},'' 
and use \Noun{Z} in selecting a candidate. 
This selection process is explained in the following section. 

Consider, for example, 
the following metonymic sentence. 

{\small
\begin{tabular}[h]{lllll}
{\it boku} & {\it ga} & {\it torusutoi} & {\it wo} & {\it yomu.}\\
(I)  & {\sf subj-case} & (Tolstoi) & {\sf obj-case} & (read)\\
\multicolumn{5}{l}{(I read Tolstoi.)}\\
\end{tabular}
}

We first gather examples containing candidate target words. 
Having already found that 
the source word is ``{\it torusutoi} (Tolstoi),'' 
we gather from the corpus all the sentences that include 
``{\it torusutoi} (Tolstoi)'' and 
transform them into syntactic structures 
by using a syntactic analyzer. 
Here, we suppose that 
sentences that include the phrase 
``{\it torusutoi no shousetsu} (Tolstoi's novel)'' 
exist. 
We extract from them 
a noun ``{\it shousetsu} (novel)'', which is 
modified by 
words ``{\it torusutoi no} (Tolstoi's)'', 
among them as a candidate of the target word. 
In this way we extract all of the \Noun{Y}s by using 
examples in the forms  
``{\it torusutoi no} \Noun{Y} (Tolstoi's \Noun{Y})'' and 
``{\it torusutoi \,} \Noun{Y} (Tolstoi \, \Noun{Y})'' 
and consider them as candidate target words. 
We also gather examples in the form of 
``\Noun{Z} {\it wo yomu} (read \Noun{Z})'' from a corpus 
for semantic restriction of the verb (\Verb{W}), 
we transform them into syntactic structures 
by using a syntactic analyzer, 
and extract all of the \Noun{Z}s 
in order to use them to define the semantic restrictions 
of the verb (\Verb{W}).

\subsubsection{Selection of the Target Word}
\label{sec:shiborikomi}

Finally, we select the target word from the 
candidates (\Noun{Y}s) 
by using the semantic restrictions of the verb (\Verb{W}) 
and the frequency of the various \Noun{Y}s. 

We first select 
those candidates satisfying the semantic restrictions 
in a verb's case frame dictionary 
because the sentence representing a metonymy interpretation 
should be literal. 
We then select from this subset of 
candidates those belonging to a set of \Noun{Z}s, 
which is extracted for the semantic restrictions of the verb (\Verb{W}) 
in the previous procedure. 
Finally, we use the frequencies of the candidate words (\Noun{Y}s) 
to select one of them. 

However, 
If this selection were based only on 
the frequency of the candidate itself, 
however, 
a candidate strongly linked to an infrequent source word 
might fail to be selected as the target word. 
To decrease the likelihood of such an error, 
we also use the frequency of the super-ordinate word\footnote{If 
X is a kind of Y, then 
Y is a {\it super-ordinate word} of X and 
X is a {\it subordinate word} of Y. 
For example, because `man' is a kind of `animal', 
`animal' is a super-ordinate word of `man' and 
`man' is a subordinate word of `animal.'}
in an {\sf is-a} hierarchy. 
When the super-ordinate word of 
a candidate word is also a candidate word, 
we weighted the frequency of each candidate word and 
added to it 
the frequency of its super-ordinate word. 
We define the frequency of the super-ordinate word of a candidate word as 
the total number of occurrences of all its super-ordinate words. 
We think this 
can compensate the frequency of a subordinate candidate word 
having a low frequency. 
In the work reported in this paper we added 
1.5 times the frequency of a candidate word to 
that of its super-ordinate word and 
selected as the desired target word 
the candidate word having the highest total frequency. 
(We have no evidence to show whether this weighting factor 
is always good. 
\footnote{
In the experiment reported in Section \ref{chap:jikken}, 
the best accuracy was obtained when the wighting factor was 1.5, 
and weighting factors of 1 and 2 each resulted in 
only one additional error. 
When the super-ordinate word frequency was not used there were 
three additional errors.})
We interpret the metonymic sentence 
by putting the target word into the input sentence. 

For an example of how this method is used, 
consider again the metonymic sentence 

{\small
\begin{tabular}[h]{lllll}
{\it boku} & {\it ga} & {\it torusutoi} & {\it wo} & {\it yomu.}\\
(I)  & {\sf subj-case} & (Tolstoi) & {\sf obj-case} & (read)\\
\multicolumn{5}{l}{(I read Tolstoi.)}\\
\end{tabular}
}

Suppose that 
the candidate target words gathered 
in the extraction 
were 
``{\it shousetsu} (a novel)-11,'' ``{\it sakuhin} (a literary product)-3,'' 
``{\it izoku} (a bereaved family)-1,'' ``{\it eikyou} (an influence)-1.'' 
(The numbers represent the frequencies.) 
First we eliminate 
candidate words such as ``{\it izoku} (a bereaved family)'' and 
``{\it eikyou} (an influence)'' 
that do not satisfy the semantic restriction of the case frame dictionary 
and do not belong to the set of \Noun{Z}s gathered using the verb (\Verb{W})\footnote{In this case, 
``{\it hon} (a book),'' ``{\it shousetsu} (a novel),'' 
``{\it sakuhin} (a literary product)'' and so on were 
gathered as \Noun{Z},  
the semantic restriction of the verb (\Verb{W}). 
So the two candidates, ``{\it shousetsu} (a novel)'' and ``{\it sakuhin} (a literary product),'' satisfied 
the semantic restriction of the verb (\Verb{W}).}. 
We then calculate each super-ordinate word frequency, 
add it to 1.5 times the frequency of its subordinate candidate word, 
and select as the target word the one 
having the highest total frequency. 
In this case, 
only ``{\it shousetsu} (a novel)'' and ``{\it sakuhin} (a literary product)'' 
have such a relationship between 
a subordinate word and a super-ordinate word. 
Therefore, the super-ordinate word frequency of 
``{\it shousetsu} (a novel)'', 
which has ``{\it sakuhin} (a literary product)'' as a super-ordinate word, 
is 3 ($ = 3 \times 1 $), and 
the other super-ordinate word frequencies are 0. 
As a result, the total frequencies  
of  ``{\it shousetsu} (a novel)'' and ``{\it sakuhin} (a literary product),'' 
which satisfy the semantic restriction, 
are respectively 19.5 ($11 \times 1.5 + 3 $) and 
4.5 ($3 \times 1.5 + 0 $). 
We select as the target word 
the word ``{\it shousetsu} (a novel)'' 
having the highest total frequency, 
and the input metonymic sentence is properly interpreted 
as ``{\it boku ga torusutoi no shousetsu wo yomu} 
(I read Tolstoi's novel). 

\section{Experiments and Discussion}\label{chap:jikken}

\subsection{Experiments}

We carried out experiments on 41 metonymic sentences 
from some textbooks (Nakamura 1977; Yamanashi 1988). 
We used as a case frame dictionary 
the NTT dictionary (Ikehara et al. 1997), 
used as a morphological analyzer JUMAN (Kurohashi and Nagao 1998), 
and used as a syntactic analyzer KNP (Kurohashi and Nagao 1994). 
Our corpus consisted of 
newspaper articles collected over a period of three years. 
We also used the NTT dictionary 
to find the relationships between 
super-ordinate and subordinate words. 
The experimental results are listed in Table \ref{tab:sougoukekka}, 
and some of the correctly interpreted metonymic sentences  
are reproduced in Table \ref{tab:kaisekirei}. 

\begin{table}[t]
\caption{Experimental results.}
\label{tab:sougoukekka}
\begin{center}
\begin{minipage}[h]{11cm}
\begin{tabular}{|@{}l@{}|@{}r@{}c@{}|}
\multicolumn{3}{l}
{$\bullet$ The sentences judged not to be metonymic sentences  (6/41)}\\
\multicolumn{3}{l}
{$\bullet$ The sentences judged to be metonymic sentences  (35/41)}\\
\hline
Correctly interpreted sentences & 66\% & (23/35) \\
\hline
Not interpreted sentences   & 17\% & (\ 6/35) \\
\noalign{\hrule height .2pt}
\hspace*{.1cm}The word did not exist in the word dictionary.  & 50\% & (3/6)\\
\hspace*{.1cm}The verb did not exist in the case frame dictionary & 50\% & (3/6)\\
\hline
Incorrectly interpreted sentences & \ 17\% & (\ 6/35)\\
\noalign{\hrule height .2pt}
\hspace*{.1cm}The target word did not exist in a corpus & 67\% & (4/6)\\
\hspace*{.1cm}Because of the frequency & 17\% & (1/6)\\
\hspace*{.1cm}Because of the wrong case frame selection & 17\% & (1/6)\\
\hline
\end{tabular}
\end{minipage}
\end{center}
\end{table}

\begin{table*}[t]
\caption{Examples of correctly interpreted metonymic sentences.}
\label{tab:kaisekirei}
\footnotesize
\begin{tabular}{|@{ }l@{}|@{ }l@{}|}
\hline
\multicolumn{1}{|c}{Metonymic Sentence}&
\multicolumn{1}{|c|}{Interpretation Sentence}\\
\hline
\begin{tabular}[h]{l@{ }l@{ }l@{ }l@{ }l}
{\it boku} & {\it ga} & \underline{\it yuumin} & {\it wo} & {\it kiku}\\
(I) & {\sf subj} &  (Yuming)  & {\sf obj} & (listen to)\\
\multicolumn{5}{l}{(I listen to \underline{Yuming}.)}\\
\end{tabular}
& 
\begin{tabular}[h]{l@{ }l@{ }l@{ }l@{ }l@{ }l@{ }l}
{\it boku} & {\it ga} & \underline{\it yuumin} & \underline{\it no} & \underline{\it uta} & {\it wo} & {\it kiku}\\
(I) & {\sf subj} &  (Yuming)  & (of) & (song) & {\sf obj} & (listen to)\\
\multicolumn{7}{l}{(I listen to \underline{the song of Yuming}.)}\\
\end{tabular}
\\[0.4cm]\hline
\begin{tabular}[h]{l@{ }l@{ }l@{ }l@{ }l}
{\it boku} & {\it ga} & \underline{\it fodo} & {\it ni} & {\it noru}\\
(I) & {\sf subj} &  (Ford)  & {\sf obj} & (drive)\\
\multicolumn{5}{l}{(I drive a \underline{Ford})}\\
\end{tabular}
&
\begin{tabular}[h]{l@{ }l@{ }l@{ }l@{ }l@{ }l@{ }l}
{\it boku} & {\it ga} & \underline{\it fodo} & \underline{\it no} & \underline{\it kuruma} & {\it ni} & {\it noru}\\
(I) & {\sf subj} &  (Ford)  & (of) & (a car) & {\sf obj} & (drive)\\
\multicolumn{5}{l}{(I drive \underline{a car of Ford})}\\
\end{tabular}
\\[0.4cm]\hline
\begin{tabular}[h]{l@{ }l@{ }l@{ }l}
\underline{\it heian jinguu} & {\it ga} & {\it mankai} & {\it da}\\
  (Heian Shrine)  & {\sf subj} & (full-bloon) & (be)\\
\multicolumn{4}{l}{(\underline{Heian Shrine} is full-bloomed.)}\\
\end{tabular}
&
\begin{tabular}[h]{l@{ }l@{ }l@{ }l@{ }l@{ }l}
\underline{\it heian jinguu} & \underline{\it no} & \underline{\it sakura} & {\it ga} & {\it mankai} & {\it da}\\
  (Heian Shrine)  & (of) & (cherry blossoms) & {\sf subj} & (full-bloom) & (be)\\
\multicolumn{6}{l}{(\underline{Cherry blossoms in Heian Shrine} are in full-bloom.)}\\
\end{tabular}
\\[0.4cm]\hline
\begin{tabular}[h]{l@{ }l@{ }l@{ }l@{ }l}
{\it boku} & {\it ga} & \underline{\it nabe} & {\it wo} & {\it taberu}\\
(I) & {\sf subj} &   (pot)  & {\sf obj} & (eat) \\
\multicolumn{5}{l}{(I eat \underline{a pot})}\\
\end{tabular}
&
\begin{tabular}[h]{l@{ }l@{ }l@{ }l@{ }l@{ }l}
{\it boku} & {\it ga} & \underline{\it nabe} & \underline{\it ryori} & {\it wo} & {\it taberu}\\
(I) & {\sf subj} &   (pot)  & (food) & {\sf obj} & (eat) \\
\multicolumn{6}{l}{(I eat \underline{hot-pot food})}\\
\end{tabular}
\\[0.4cm]\hline
\begin{tabular}[h]{l@{ }l@{ }l@{ }l@{ }l}
\underline{\it shiro-bai} & {\it ga} & {\it ihansha} & {\it wo} \\
  (a white motorbike)  & {\sf subj} & (a lawbreaker) & {\sf obj}\\
{\it taiho-su}\\
(arrest)\\
\multicolumn{5}{l}{(\underline{A white motorbike} arrests a lawbreaker.)}\\
\end{tabular}
&
\begin{tabular}[h]{l@{ }l@{ }l@{ }l@{ }l@{ }l}
\underline{\it shiro-bai} & \underline{\it keikan} & {\it ga} & {\it ihansha} & {\it wo} \\
  {\footnotesize (a white motorbike)} & {\footnotesize (a policeman)}  & {\footnotesize {\sf subj}} & {\footnotesize (a lawbreaker)} & {\footnotesize {\sf obj}} \\
 {\it taiho-su}\\
 {\footnotesize (arrest)}\\
\multicolumn{6}{l}{(\underline{A white motorbike policeman} arrests a lawbreaker.)}\\
\end{tabular}
\\[0.4cm]\hline
\end{tabular}
\end{table*}

\subsection{Discussion}

In the experiment, 
six of the 41 input sentences were judged 
not to be metonymic sentences. 
This was because 
each case element of these sentences satisfied 
the semantic restrictions of the case frame dictionary. 
Although these sentences,
such as ``{\it boku ga \underline{niwa} wo haku.} 
(I sweep \underline{the garden}.),'' 
are classified as metonymic sentences in textbooks, 
they are generally thought to be interpreted literally. 
We correctly interpreted 23 of the remaining 35 sentences, 
including the interesting sentence 
``\underline{\it heian jinguu} {\it ga} {\it mankai} {\it da} (\underline{Heian Shrine} is in full-bloom)'' interpreted as 
``\underline{\it heian jinguu no sakura}  {\it ga}  {\it mankai}  {\it da} (\underline{Cherry blossoms in Heian Shrine} are in full-bloom).'' 
(Heian Shrine is in Kyoto, the beautiful ancient capital in Japan. 
The cherry blossoms are lovely in spring. ) 

Our method correctly 
interpreted 23 of the 41 input sentences. 
Among the 41 input sentences treated as metonymy expressions in 
textbooks, however, 6 could be interpreted literally. 
Thus, our method correctly analyzed 23 cases 
of 35 metonymic expressions (66\%) 
that could not be analyzed correctly by the former NLP systems. 

Next, we examine problems clarified 
in the experiment. 

There were some cases of misinterpretation 
because the input sentence included 
a verb or a noun, that was not 
in the case frame dictionary or noun dictionary, 
such as ``gennari suru (be fed up)'' and ``Yoshimoto Banana.'' 
And the sentence ``{\small \it boku ga} 
{\small \it \underline{isshou bin}} 
\underline{({\it no sake})} {\it  wo nomu.} 
(I drink \underline{a one-{\it sho} bottle (of liquor)})'' 
was misinterpretated 
because there were few examples 
and the correct target word was not among them. 
The examples gathered in the analysis of this sentence were 
``{\it rappanomi} (drinking from a bottle)-1,'' 
``{\it raberu} (label)-1,'' and 
``{\it sen} (a cap)-1.'' 
The correct target word, ``{\it sake} (liquor),'' 
was not among them. 
This problem can be overcome 
by enlarging the corpus from which the examples are extracted. 

\section{Conclusion}\label{chap:keturon}

This paper described a method for interpreting 
metonymic sentences by using examples. 
A metonymic sentence is considered 
a sentence from which a target word is absent. 
Because the Japanese particle ``{\it no} (of)'' can be 
used to express 
almost all possible relationships 
between a metonym (source word) and the omitted target word, 
we interpret a metonymic sentence 
by using examples in the form ``\Noun{X} {\it no} \Noun{Y}'' 
to identify 
the omitted target word rather than by 
using a hand-built database on knowledge of metonymy 
to identify it. 
Specifically, 
\begin{enumerate}
\item We specify a source word (``\Noun{X}'') by 
using a case frame dictionary. 
\item 
We gather from a corpus examples 
in the forms of ``\Noun{X} {\it no} \Noun{Y} (\Noun{Y} of \Noun{X})'' and ``\Noun{X} \Noun{Y}'' 
that include ``\Noun{X}'' (the source word), 
and from these examples we get multiple \Noun{Y}s 
as candidate target words. 
\item 
We select one candidate as the desired target word 
by using the 
semantic restriction of the verb (\Verb{W}) and the frequency of \Noun{Y}. 
\end{enumerate}

Evaluating this method experimentally, 
we found that it correctly interpreted 
23 of 35 metonymic sentences (66\%). 

Two advantages of this method are 
that a hand-built metonymy database is not necessary and 
that we will be able to interpret newly-coined metonymic sentences 
by using a new corpus. 

It is unfortunately necessary 
to extract metonymic sentences by using some other system, 
because this method cannot distinguish metonymic sentences 
from metaphorical sentences and literal sentences. 
And this method cannot interpret the metonymic sentence properly, 
when the corpus contains few examples that include the source word. 

\section*{Acknowledgment}
We would like to express our sincere appreciation to 
the people who allowed us to use 
the NTT dictionary ``{\it Nihongo Goi Taikei} (Japanese Lexical Dictionary)'' (Ikehara et al. 1997) in Kyoto University.

\end{document}